%% file: main.tex
\documentclass[10pt,twocolumn,letterpaper]{article}

\usepackage{cvpr}              
\input{preamble}
\definecolor{cvprblue}{rgb}{0.21,0.49,0.74}
\usepackage[pagebackref,breaklinks,colorlinks,allcolors=cvprblue]{hyperref}
\usepackage[accsupp]{axessibility}  
\usepackage{algorithm}
\usepackage{algorithmic}
\usepackage{marvosym}

\newcommand{\q}[1]{\textbf{#1}}
\newcommand{\U}{$\uparrow$}
\newcommand{\D}{$\downarrow$}
\newcommand{\clr}[1]{{\color{red}#1}}
\newcommand{\cly}[1]{{\color{yellow}#1}}

\newcommand{\settablefontnine}{\fontsize{9}{11.5}\selectfont}

\def\paperID{} 
\def\confName{CVPR}
\def\confYear{2026}

\title{VAD-GS: Visibility-Aware Densification for 3D Gaussian Splatting\\in Dynamic Urban Scenes}

\author{Yikang Zhang\textsuperscript{1}\\
{\tt\small yikangzhang@tongji.edu.cn}
\and
Rui Fan\textsuperscript{1,2,3}\textsuperscript{\Letter}\\
{\tt\small rui.fan@ieee.org}
\and 
\textsuperscript{1}\fontsize{11.3pt}{13pt}\selectfont Shanghai Research Institute for Intelligent Autonomous Systems, Tongji University \\
\textsuperscript{2}\fontsize{11.3pt}{13pt}\selectfont College of Electronic and Information Engineering, Tongji University \\
\textsuperscript{3}\fontsize{11.3pt}{13pt}\selectfont National Key Laboratory of Human-Machine Hybrid Augmented Intelligence, Xi’an Jiaotong University\\
}

\begin{document}
\maketitle
\begingroup
\renewcommand\thefootnote{}
\footnotetext{\textsuperscript{\Letter}Corresponding author.}
\endgroup

\input{sec/0_abstract}    
\input{sec/1_intro}
\input{sec/2_related_work}
\input{sec/3_preliminaries}
\input{sec/4_method}

\input{sec/5_experiments}

\input{sec/6_ablation}
\input{sec/7_conclusion}
\input{sec/8_acknowledge}
{
    \small
    \bibliographystyle{unsrt}
    \bibliography{main}
}
\input{sec/X_suppl}


\end{document}

%% file: sec/0_abstract.tex
\begin{abstract}
3D Gaussian splatting (3DGS) has demonstrated impressive performance in synthesizing high-fidelity novel views. Nonetheless, its effectiveness critically depends on the quality of the initialized point cloud. Specifically, achieving uniform and complete point coverage over the underlying scene structure requires overlapping observation frustums, an assumption that is often violated in unbounded, dynamic urban environments. Training Gaussian models with partially initialized point clouds often leads to distortions and artifacts, as camera rays may fail to intersect valid surfaces, resulting in incorrect gradient propagation to Gaussian primitives associated with occluded or invisible geometry. Additionally, existing densification strategies simply clone and split Gaussian primitives from existing ones, incapable of reconstructing geometry from missing structures. To address these limitations, we propose VAD-GS, a 3DGS framework tailored for geometry recovery in challenging urban scenes. Our method identifies unreliable geometry structures via voxel-based visibility reasoning, selects informative supporting views through diversity-aware view selection, and recovers missing structures via multi-view stereo reconstruction. This design enables the generation of new Gaussian primitives guided by reliable geometric priors, even in regions lacking initial points. Extensive experiments on the Waymo and nuScenes datasets demonstrate that VAD-GS outperforms state-of-the-art 3DGS approaches and significantly improves the quality of reconstructed geometry for both static and dynamic objects. Our project webpage is at \url{mias.group/VAD-GS/}.
\end{abstract}
\vspace{-1em}
\begin{figure}[t!]
\centering
\includegraphics[width=1\columnwidth]{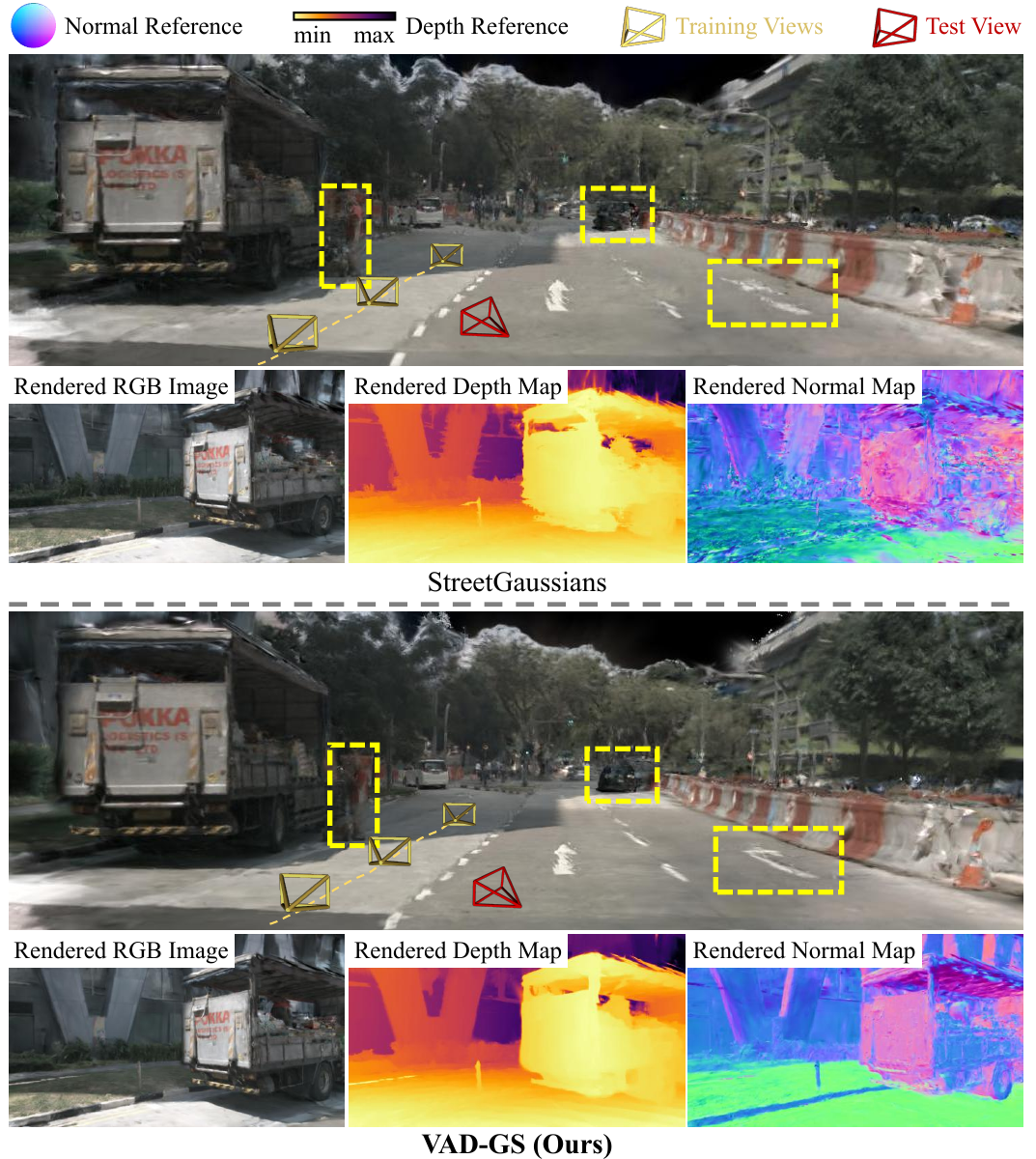} 
\caption{
\textbf{A comparison between VAD-GS and StreetGaussians.} 
While both methods achieve comparable rendering quality, VAD-GS demonstrates superior recovery of incomplete or unreliable scene geometry, as evidenced by notable improvements in the rendered depth and normal maps. The supplement provides additional results, including a video demonstration of VAD-GS.
}
\label{fig1.comparison}
\end{figure}

%% file: sec/1_intro.tex
\section{Introduction}
\label{sec:intro}

Realistic simulation is critical for the development and validation of autonomous driving systems \cite{3dgs_survey}. Traditional simulators rely on handcrafted assets, inherently limiting scene scalability and diversity \cite{dosovitskiy2017carla}. Recent advances in neural scene representations have enabled data-driven, photorealistic novel view synthesis (NVS), which provides a more efficient and scalable alternative. Specifically, neural radiance field (NeRF)-based approaches \cite{mildenhall2021nerf} represent scenes using neural networks and achieve high-fidelity 3D reconstruction. Nevertheless, volume rendering is typically computationally intensive, thereby limiting the practical applicability of NeRF and its variants. To enable real-time rendering, 3D Gaussian splatting (3DGS) explicitly represents scenes as anisotropic 3D Gaussian primitives with learnable geometry and appearance attributes. These primitives are jointly optimized to align with the underlying scene structure by enforcing photometric consistency across images captured from multiple views. Building on its impressive performance in object-level reconstruction, recent extensions 
\cite{kerbl2024hierarchical, 4dgs} have demonstrated strong potential for reconstructing complex urban scenes.

Despite these advances, recovering complete and reliable geometry for unbounded environments from sparse observations remains a major challenge. In 3DGS, scene completeness and texture details are typically enhanced by splitting or cloning existing Gaussian primitives, which are initialized from point clouds obtained through structure from motion (SfM) or LiDAR scan accumulation \cite{3dgs_survey}. However, autonomous driving datasets present inherent limitations: (1) Unlike scene-centric reconstruction settings, multiple cameras mounted on a vehicle capture outward-facing views in urban scenes with limited overlap (typically less than 15\%) \cite{wei2024omniscene}, which makes stereo matching between adjacent, synchronous images unreliable; (2) Although multi-view stereo (MVS) methods can recover static scene structure from asynchronous video frames, they are ineffective in reconstructing dynamic objects; (3) Despite the incorporation of LiDAR points to enhance geometric consistency in pioneering studies \cite{yan2024street, zhou2024drivinggaussian}, substantial blind spots in scene structure persist due to the limited field of view. 
Consider a typical scenario where a traffic sign is positioned beyond the LiDAR’s vertical range (as shown in the supplement and video). Its low-texture surface also provides insufficient visual cues for reliable SfM reconstruction across images. As a result, the initialized point cloud fails to recover accurate surface geometry. 
Furthermore, during Gaussian training, photometric errors caused by missing geometry are erroneously attributed to background structures, such as trees or buildings behind the sign. Consequently, gradient-based splitting and cloning operations may inadvertently be applied to invisible Gaussian primitives. Although this training process may improve rendering quality for specified views, it distorts the underlying scene geometry and ultimately degrades generalization to unseen perspectives.

Several recent studies have focused on enhancing scene completeness within the original 3DGS framework. For instance, GeoTexDensifier \cite{jiang2024geotexdensifier} incorporates additional depth and normal priors to guide the splitting process for improved surface alignment, whereas DNGaussian \cite{li2024dngaussian} identifies missing geometry by performing global-local normalization between the rendered depth and that estimated using DPT \cite{ranftl2021vision}. Nonetheless, these methods are confined to regions with existing Gaussian primitives and are incapable of handling uninitialized areas. To overcome this drawback, GaussianPro \cite{cheng2024gaussianpro} introduces a patch matching-based geometry completion strategy, which leverages stereo constraints from a set of images with precomputed camera poses to generate additional point clouds independent of the rendering process. While GaussianPro greatly enhances geometry recovery, it remains limited to static scenes and is incapable of handling dynamic object densification. Moreover, it relies solely on adjacent frames captured using a single camera, thereby missing long-range temporal dependencies and cross-camera visual cues.

To address the aforementioned challenges, this paper introduces a {\textbf{v}isibility-\textbf{a}ware \textbf{d}ensification framework for 3D \textbf{G}aussian \textbf{s}platting (\textbf{VAD-GS})} tailored for dynamic, unbounded urban environments. Unlike previous approaches that passively react to photometric errors, VAD-GS actively evaluates structural completeness and selectively reconstructs incomplete regions by leveraging views that provide the most reliable stereo geometry. Specifically, we introduce a voxel-based object surface visibility reasoning approach that provides geometric priors for both static backgrounds and dynamic objects. Each voxel aggregates view-dependent visibility information of the corresponding 3D points, thereby enabling occlusion-aware modeling through depth rasterization with z-buffering. Furthermore, we propose a diversity-aware view sampling strategy that selects informative supporting views for each reference view, aiming to balance view frustum overlap and triangulation quality. The selected views are processed using a patch matching-based MVS algorithm to extract depth and normal information, which serves as reliable geometric priors for new Gaussian primitive initialization and scene consistency enforcement. VAD-GS is evaluated on the Waymo Open dataset \cite{waymo} and the nuScenes dataset \cite{caesar2020nuscenes}, both of which contain complex urban dynamics and sparse multi-view observations. Extensive experiments demonstrate that VAD-GS achieves state-of-the-art (SoTA) rendering quality while producing more consistent geometry with fewer artifacts compared to previous SoTA methods (see Fig. \ref{fig1.comparison}). The main contributions of this study are summarized as follows:
\begin{itemize}
    \item A novel Gaussian splatting framework tailored for dynamic urban scenes, which actively completes missing geometry using multi-camera, cross-frame observations.
    \item A voxel-based surface visibility reasoning approach that identifies unreliable static and dynamic object geometry.
    \item A diversity-aware sampling strategy that improves MVS reconstruction quality by optimizing supporting views.
    \item An extension of MVS reconstruction to dynamic, multi-camera driving scenarios, enabling both Gaussian densification and scene consistency enforcement.
\end{itemize}

\begin{figure*}[t!]
\centering
\includegraphics[width=1\textwidth]{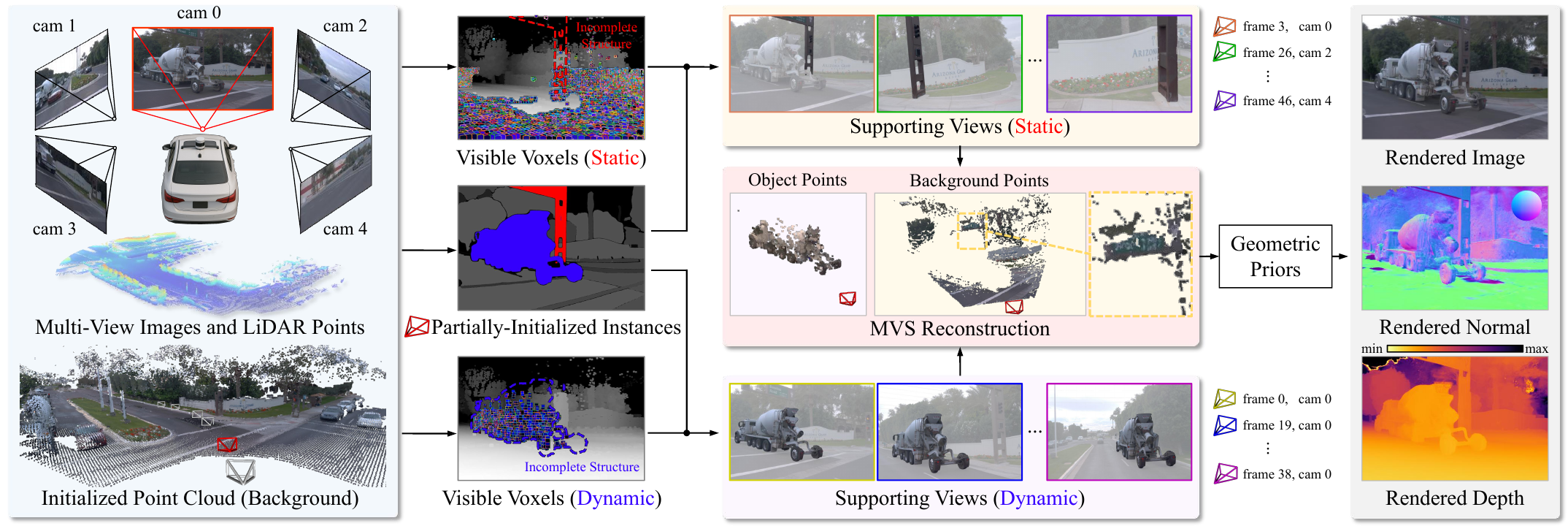} 
\caption{
\textbf{VAD-GS pipeline.} 
For each static or dynamic instance with incomplete geometry, VAD-GS first performs voxel-based visibility reasoning to identify a set of potential observation views. It then incrementally selects diverse supporting views to perform MVS reconstruction. The resulting geometric priors are subsequently used for Gaussian densification and optimization.
}
\label{fig2.flowchart}
\vspace{-3mm}
\end{figure*}

%% file: sec/2_related_work.tex
\section{Related Work}
\subsection{Novel View Synthesis}
NVS aims to generate photorealistic images of objects or scenes from previously unseen viewpoints, without explicitly modeling 3D simulation assets or lighting maps. NeRF \cite{mildenhall2021nerf}, a pioneering work in this field, represents 3D scenes as learnable volumetric density fields, parameterized by a large multi-layer perceptron (MLP). Subsequent studies have primarily focused on improving both rendering quality and computational efficiency.
For example, Instant-NGP \cite{instant-ngp} introduces a multi-resolution hash encoding scheme that adaptively allocates higher representational capacity to geometrically complex regions, thereby significantly improving rendering efficiency. 
Mip-NeRF \cite{barron2021mip} improves point sampling in ray marching to mitigate aliasing artifacts caused by resolution mismatches, while its extension Mip-NeRF 360 \cite{mipnerf360} further adapts the approach to handle unbounded scenes.
However, the computational cost of volume rendering remains a major barrier to the practical deployment of NeRF models in real-world applications.

Recent 3DGS approaches \cite{3dgs} have introduced an alternative NVS paradigm, enabling real-time rendering of large-scale scenes. By projecting anisotropic 3D Gaussian ellipsoids onto the 2D image plane using splatting-based rasterization and computing pixel colors through depth sorting and $\alpha$-blending, these methods effectively circumvent the computational overhead of ray marching. Since then, several studies have extended 3DGS to dynamic urban scenes. Notably, StreetGaussians \cite{yan2024street} models dynamic vehicles in the foreground as rigid groups of Gaussian primitives and employs a 4D spherical harmonics model to capture appearance variation over time. Similarly, DrivingGaussian \cite{zhou2024drivinggaussian} uses a composite dynamic Gaussian graph to model multiple dynamic objects, OmniRe \cite{omnire} incorporates skinned multi-person linear Gaussians to represent non-rigid entities such as pedestrians and cyclists, while HUGSIM \cite{zhou2025hugsim} introduces a multi-plane ground model to alleviate lane distortion in extrapolated views. Other works \cite{song2025coda, tourani2025leveraging, huang2023neural, li2025anchored} have also explored related extensions in different directions.
Despite the appealing results achieved by this paradigm, two major limitations persist: (1) new Gaussian primitives are typically generated by splitting or cloning existing ones, making reconstruction quality highly dependent on the accuracy and completeness of the initialized point clouds; (2) recovering missing geometry based solely on photometric errors is inherently challenging, as gradient updates may be incorrectly propagated to view-proximal yet geometrically unrelated Gaussians, leading to the distortion of neighboring Gaussians and ultimately degrading the overall scene geometry. To address these issues, we explore the incorporation of additional geometric cues, particularly MVS constraints, to obtain more reliable structural information beyond gradient propagation.

\subsection{Multi-View Stereo}
MVS is a fundamental computer vision technique that reconstructs dense 3D scene geometry from a set of images with known intrinsic parameters \cite{mvs_survey}. Online MVS methods typically select keyframes from low-resolution video streams to perform real-time camera pose estimation and point cloud generation \cite{scannet}. In contrast, offline MVS approaches, typically following the SfM pipeline \cite{schonberger2016structure, DTU, ETH3d}, aim for high-resolution, large-scale scene reconstruction at the expense of greater computational complexity. Despite their differing objectives, both pipelines face critical challenges that directly impact reconstruction quality, particularly in view selection and depth estimation. For view selection, GP-MVS \cite{gpmvs} employs a heuristic pose-distance measure function to select informative keyframes, while MVSNet \cite{yao2018mvsnet} introduces a score function that ranks neighboring views based on frustum overlap. In terms of depth estimation, plane sweeping-based methods discretize depth candidates to construct cost volumes and measure feature similarity across warped views, thereby favoring reconstructions with higher resolution \cite{cheng2020deep, yang2022non, guo2024planestereo}. In contrast, patch matching-based approaches achieve high efficiency by randomly sampling depth guesses for individual pixels and iteratively propagating plausible estimates from neighboring pixels \cite{Xu2020ACMP, wang2021patchmatchnet}. In this work, we exploit MVS consistency not only to guide the densification of Gaussian primitives for each object, but also to provide geometric supervision that complements photometric gradient-based optimization. More importantly, we extend MVS consistency to dynamic settings where both objects and cameras are in motion, which remains challenging for existing methods.

%% file: sec/3_preliminaries.tex
\section{Preliminaries}
3DGS-based approaches represent a scene using a set of anisotropic 3D Gaussian primitives \cite{3dgs}. Each primitive $\boldsymbol{x}$ is modeled by a Gaussian distribution, defined as follows:
\begin{equation}
    G(\boldsymbol{x}) = \exp\left(-\frac{1}{2}(\boldsymbol{x} - \boldsymbol{\mu})^{\top}\boldsymbol{\Sigma}^{-1}(\boldsymbol{x} - \boldsymbol{\mu})\right),
\end{equation}
where $\boldsymbol{\mu} \in \mathbb{R}^3$ denotes the primitive center, and the covariance matrix $\boldsymbol{\Sigma}$ is parameterized using a scaling factor and a rotation quaternion. Unlike volumetric representations such as NeRF, 3DGS avoids the computational overhead of volumetric ray marching by adopting a tile-based rasterization pipeline. The color $C$ at each pixel is obtained by compositing the overlapping Gaussians along the pixel ray via front-to-back $\alpha$-blending, as expressed as follows:
\begin{equation}
    C = \sum_{i \in \mathcal{N}} \boldsymbol{c}_i \alpha_i \prod_{j=1}^{i-1}(1 - \alpha_j),
\end{equation}
where $\mathcal{N}$ denotes the set of Gaussian primitives intersected by the pixel ray, while $\boldsymbol{c}_i$ and $\alpha_i$ represent the color and opacity of the $i$-th primitive, respectively. 
The photometric error of $C$ propagates gradients to Gaussians that exhibit inconsistencies in shape or appearance. To improve scene geometry representation, 3DGS adopts a densification strategy by splitting existing Gaussians with large covariance into smaller primitives. Theoretically, these newly generated Gaussians are iteratively refined to converge toward consistent geometry and appearance, thereby improving both coverage and structural fidelity.

%% file: sec/4_method.tex
\begin{figure}[t!]
\centering
\includegraphics[width=1\columnwidth]{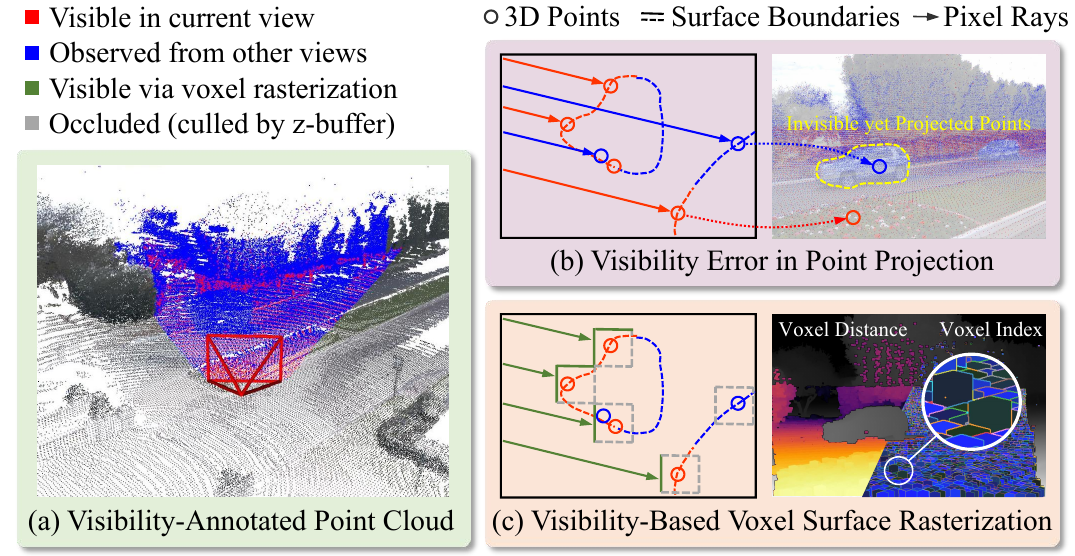} 
\caption{
\textbf{Voxel-based visibility reasoning.} 
(a) Red points are visible, whereas blue points, captured from other views, are invisible in the reference view. (b) The invisibility of blue points may result from occlusions or insufficient sampling rays in the reference view.
(c) Rasterizing the distances and indices of visible voxels (in green) yields dense depth maps and accurate pixel-voxel mapping. 
}
\label{fig3.visibility}
\vspace{-3mm}
\end{figure}

\section{Methodology}
As illustrated in Fig. \ref{fig2.flowchart}, the proposed VAD-GS framework identifies visible surfaces of unreliable geometry, selects informative supporting views, and complements missing structures via MVS-guided densification.

\subsection{Voxel-Based Visibility Reasoning}\label{method.discriminant} 
The effectiveness of 3DGS depends critically on well-initialized point clouds.
In the training phase, photometric-guided optimization and densification aggregate all Gaussian primitives along each viewing ray via alpha compositing. However, when the initialized geometry is incomplete, photometric error gradients may erroneously accumulate on Gaussians associated with occluded or invisible surfaces, leading to mismatched appearance, geometric distortions, and floater artifacts \cite{zhang2024pixelgs}.
To address this issue, explicit view-dependent visibility and occlusion reasoning for scene objects must be incorporated into the reconstruction process. As illustrated in Fig. \ref{fig3.visibility}, a 3D point sampled along a pixel ray should correspond to the first intersected surface, visible from the given viewpoint. Existing methods either extract visible points independently from each single view or aggregate points from all available views without considering occlusion. While sampling points from a specific view guarantees valid surface intersections, it provides limited scene coverage due to observation constraints. In contrast, aggregating multiple point clouds from different timestamps improves spatial coverage but lacks occlusion awareness, allowing rays to traverse occluded structures and resulting in erroneous updates on non-visible geometry.

To enable efficient reasoning about scene visibility and occlusion, in the first step, VAD-GS applies voxelization to the initialized point cloud to enforce uniform spatial density. The visibility of each voxel is defined as the union of the observation views associated with its constituent points. Specifically, each LiDAR point is sourced from an individual frame, whereas SfM points are triangulated from at least two views. Reasoning about voxel visibility provides two key advantages: (1) Rasterizing the distances of visible voxel surfaces via classical z-buffering \cite{akenine2018realtime} produces a denser and more reliable depth map, compared to conventional geometric supervision methods that rely on sparse point clouds and nearest-neighbor search. This rasterized depth map reduces missed surface intersections, constrains depth errors within the voxel resolution, and naturally excludes occluded voxels located behind incomplete foreground geometry, thereby preventing erroneous updates during model optimization. (2) Rasterizing a 2D index map establishes a mapping between image pixels and their underlying 3D structures. By storing only the index of the first intersected and visible voxel along each pixel ray, this mapping ensures both validity and efficiency, enabling fast retrieval of geometric attributes such as 3D position, surface normals, and neighborhood connectivity.

Voxel visibility is further utilized to identify incomplete scene structures. Specifically, scene elements such as vehicles, trees, and buildings are individually extracted using an offline instance segmentation network \cite{sam}. Pixels belonging to each segmented instance are then mapped to their corresponding voxels according to the rasterized index map. For each instance, two depth values are compared: one rasterized from visible voxels via z-buffering, while the other rendered from existing Gaussians through the splatting pipeline. If the depth rendered from Gaussians is consistently smaller than the voxel-derived depth, it indicates either successful completion of previously missing geometry or acceptable redundancy, both of which can be effectively handled via opacity adjustment. In contrast, if the Gaussian-rendered depth is absent or significantly larger than the voxel depth, it implies that the geometry is either incompletely initialized or distorted in earlier optimization stages. In such cases, the instance is flagged for re-initialization to improve reconstruction completeness.

\subsection{Diversity-Aware View Selection} \label{method.view_selection}
After identifying instances with unreliable structures, the corresponding voxels along with available views can be retrieved using the rasterized index map. To ensure reliable geometry reconstruction via MVS, it is essential to select a representative subset of views that provides strong geometric constraints. Although consecutive video frames captured by a single camera can provide sufficient overlap for 3D reconstruction in static scenes \cite{cheng2024gaussianpro}, their performance deteriorates in driving scenarios due to the existence of both dynamic objects and continuous ego-motion. These factors necessitate sufficient frustum overlap and strong stereo constraints, which can only be achieved by selecting views from different cameras and timestamps. To this end, we define the following score:
\begin{equation} \label{eq.diversity}
    s = \frac{N}{ \boldsymbol{d}_{R} ^\top \boldsymbol{d}_{S} } \frac{\sqrt{{t_x}^2 + {t_y}^2} }{|t_z| + \epsilon}  \sin \theta,
\end{equation}
to quantify the geometric diversity between a pair of views, where $\boldsymbol{d}_{R}$ and $\boldsymbol{d}_{S}$ are column vectors that store the distances from $N$ voxels visible in both reference and supporting views to their respective viewpoints, $\boldsymbol{t}= (t_x, t_y, t_z)^\top$ denotes the relative translation between the two views, $\epsilon$ ensures numerical stability, and $\theta$ represents the angular difference between their orientations. Higher scores are achieved when voxels are denser and closer to both views, lateral variations are greater, longitudinal displacements are minimal, and orientation differences are larger. In contrast, lower scores result under opposite conditions, implying fewer stereo cues. Diverse supporting views selected based on this score are subsequently utilized for MVS reconstruction.
To balance information diversity and computational cost, a top-$k$ subset is selected for each reference view via \eqref{eq.diversity}. Pairwise diversity among the supporting views is also considered to avoid overly similar viewpoints.

\begin{figure}[t!]
\centering
\includegraphics[width=0.98\columnwidth]{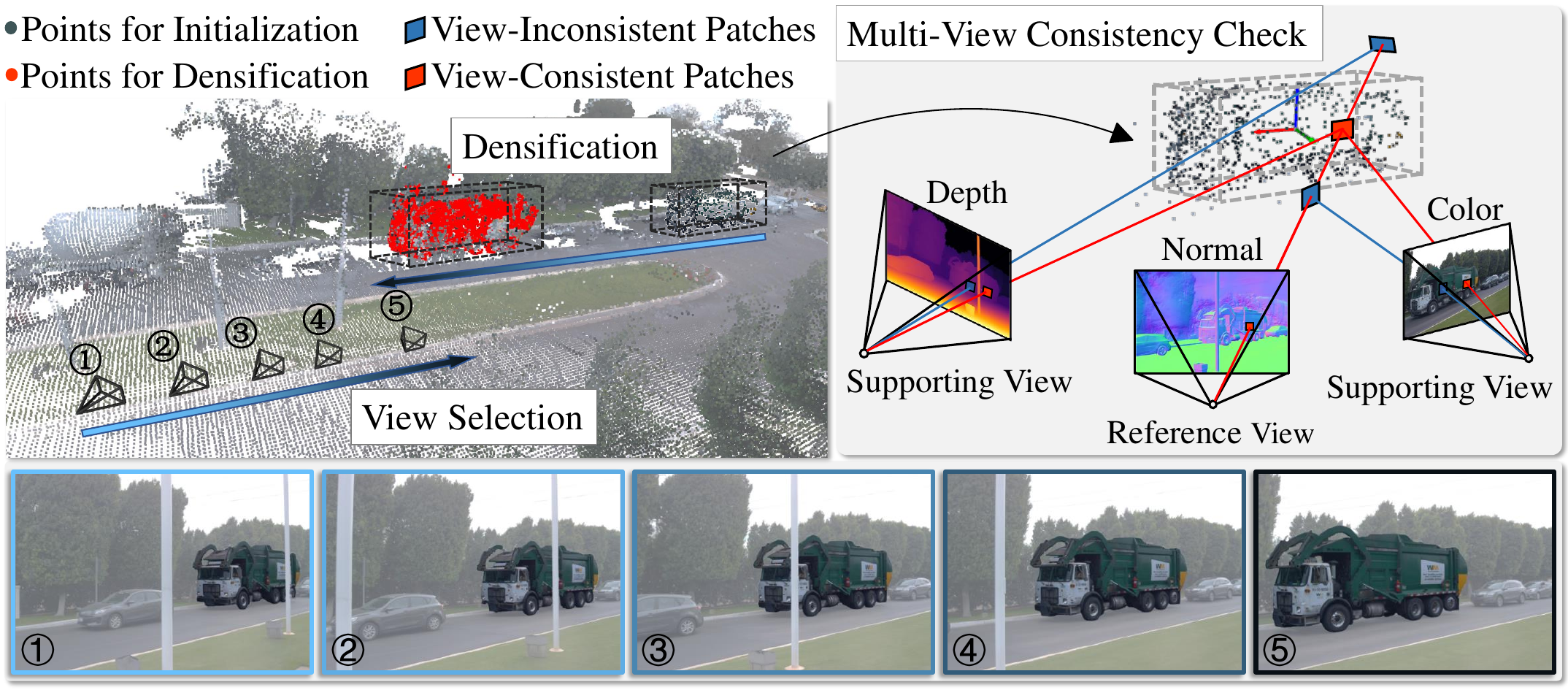} 
\caption{
\textbf{View Selection and MVS Reconstruction}. 
Image patches are warped across views to check the consistency of depth, normal, and color. Only consistently matched patches (in red) are considered valid for MVS reconstruction, while inconsistent ones (in blue) are discarded. The reconstructed geometry is then used to guide Gaussian densification.
}
\label{fig4.patch_matching}
\vspace{-3mm}
\end{figure}

\subsection{MVS Reconstruction from Selected Views} \label{method.pmvs}
By incorporating the selected supporting views that provide sufficient geometric constraints, reliable 3D points can be generated to complement incomplete scene structures. To achieve this goal, we adopt a multi-view patch matching approach, which has been widely used for dense 3D reconstruction in static scenes \cite{Xu2020ACMP}. The method estimates local surface planes by matching small image patches across multiple views, under the assumption that the scene geometry is locally piecewise planar. 
Specifically, an image pixel located at $\boldsymbol{p}=(u,v)^\top$ is associated with a local 3D plane, expressed as: $z \, \boldsymbol{n}^\top \boldsymbol{K}^{-1} \widetilde{\boldsymbol{p}}+d=0$, where $z$ denotes the depth value at $\boldsymbol{p}$, $\boldsymbol{n}$ represents the corresponding surface normal, $\boldsymbol{K}$ denotes the camera intrinsic matrix, $\widetilde{\boldsymbol{p}}$ represents the homogeneous coordinates of $\boldsymbol{p}$, and $d$ represents the distance between the surface and the camera origin. 
In the patch matching process, $\boldsymbol{p}$ in the reference view, associated with a plane hypothesis $(d,\boldsymbol{n})$, is projected to $\boldsymbol{p}'$ in a supporting view using the following expression: 
\begin{equation}
\widetilde{\boldsymbol{p}}' \simeq \boldsymbol{K} \left( \boldsymbol{R} - \frac{\boldsymbol{t} \boldsymbol{n}^\top}{d} \right) \boldsymbol{K}^{-1} \widetilde{\boldsymbol{p}},
\end{equation}
where $[\boldsymbol{R}, \boldsymbol{t}]$ denotes the relative pose from the reference view to the supporting view. Patches are warped across views to assess photometric consistency using RGB images and geometric consistency using depth and normal maps. As shown in Fig. \ref{fig4.patch_matching}, a pair of patches is deemed consistent if the reference patch and warped supporting patch exhibit similar image features, and their plane hypotheses $(d,\boldsymbol{n})$ are well aligned. 
These hypotheses are iteratively refined by propagating candidates from neighboring pixels under the assumption of local structural similarity. Gaussian-rendered results serve as reasonable initial guesses, effectively reducing the number of random re-sampling iterations required when consistent matches are absent.
Through repeated updates and consistency checks, patches consistent across the majority of views are obtained for robust reconstruction.

Although reliable static structures can be recovered in this way \cite{Xu2020ACMP, cheng2024gaussianpro}, handling dynamic objects remains a significant challenge. 
In theory, a rigid vehicle in motion can be treated as static by transforming all observation views into its local coordinate system. However, in practice, object masks derived from 3D bounding boxes cannot accurately delineate the boundaries between foreground and background.
Including pixels from the static background may introduce misleading patch matches and disrupt the local geometric consistency assumptions for the moving object. Although instance segmentation methods provide more precise, contour-aligned masks, their performance is highly sensitive to the quality of input prompts. A prompt derived from a single LiDAR frame may fail to capture the entire object due to limited point coverage, whereas one generated from a temporally aggregated point cloud often introduces occluding scene structures unrelated to the object.
This challenge, nevertheless, can be effectively addressed by leveraging our rasterized visible voxels, which inherently incorporate occlusion cues and provide more accurate instance-level prompts for segmentation. Consequently, the method restricts the patch matching process within either static or dynamic regions across all relevant views, effectively minimizing cross-region interference and greatly enhancing reconstruction accuracy.

\subsection{Loss Function}
We optimize the model by minimizing a weighted sum of four loss terms, as expressed as follows:
\begin{equation} \label{eq.loss}
    \mathcal{L} = \mathcal{L}_{\text{color}} + \lambda_\text{normal} \mathcal{L}_{\text{normal}} + \lambda_\text{hard} \mathcal{L}_{\text{hard}} + \lambda_\text{soft} \mathcal{L}_{\text{soft}},
\end{equation}
where $\mathcal{L}_{\text{color}}$ quantifies the discrepancy between rendered and observed images, $\mathcal{L}_{\text{normal}}$ quantifies the angular deviations between the rendered surface normals and those obtained via patch matching, 
and $\mathcal{L}_{\text{hard}}$ and $\mathcal{L}_{\text{soft}}$ measure depth errors using fixed and learned Gaussian opacities, respectively. 
By incorporating our visibility-aware Gaussian densification strategy, the optimization of \eqref{eq.loss} achieves superior geometric reconstruction performance in complex dynamic urban settings.

%% file: sec/5_experiments.tex
\section{Experiments}

\begin{table}[t!]
\renewcommand\arraystretch{0.9}
\setlength\tabcolsep{5pt} 
\centering
\settablefontnine
\begin{tabular}{l c c c c}
\toprule
                                & PSNR$\uparrow$    & PSNR*$\uparrow$   & SSIM$\uparrow$    & LPIPS$\downarrow$ \\
\hline
\hline
3D GS \cite{3dgs}               & 29.64             & 21.25             & 0.918             & 0.117 \\
NSG \cite{ost2021neural}        & 28.31             & 24.32             & 0.862             & 0.346 \\
MARS \cite{wu2023mars}          & 29.75             & 26.54             & 0.886             & 0.264 \\
EmerNeRF \cite{yang2023emernerf}& 30.87             & 21.67             & 0.905             & 0.133 \\
PVG \cite{pvg}                   & 31.82             & 24.68             & 0.910             & 0.122 \\
OmniRe \cite{omnire}             & 31.12             & 25.20             & 0.902             & 0.123 \\
StreetGS \cite{yan2024street}   & 34.61             & 30.23             & 0.938             & 0.079 \\
Ours                            & \textbf{35.59}    & \textbf{31.31}    & \textbf{0.950}    & \textbf{0.047} \\ 

\bottomrule
\end{tabular}
\caption{Quantitative results on Waymo Open Dataset. PSNR*: evaluated on dynamic objects only.}
\label{table1.waymo}
\vspace{-3mm}
\end{table}

\begin{table*}[t!]
\centering
\settablefontnine
\renewcommand\arraystretch{0.9}
\setlength\tabcolsep{2.3pt} 
\begin{tabular}{l cccc cccc cccc cccc}
\toprule
~        & \multicolumn{4}{c}{PVG \cite{pvg} } & \multicolumn{4}{c}{OmniRe \cite{omnire}} & \multicolumn{4}{c}{StreetGS \cite{yan2024street}} & \multicolumn{4}{c}{\textbf{VAD-GS (Ours)}} \\ 
           \cmidrule(r){2-5}          \cmidrule(r){6-9}            \cmidrule(r){10-13}                   \cmidrule(r){14-17} 
~        &  PSNR\U & SSIM\U & LPIPS\D &\#G  & PSNR\U  & SSIM\U & LPIPS\D & \#G &  PSNR\U & SSIM\U & LPIPS\D &\#G  &  PSNR\U & SSIM\U & LPIPS\D &\#G \\ 
\hline
\hline
Scene 00  &   22.77 &   0.63 &   0.22 &195k &   22.71 &   0.63 &   0.16 &153k &   22.87 &   0.65 &\q{0.13}&146k &\q{25.54}&\q{0.81}&   0.16 &234k \\ 
Scene 01  &   22.91 &   0.66 &   0.24 &196k &   22.27 &   0.64 &\q{0.22}&108k &   22.12 &   0.66 &   0.23 &109k &\q{23.69}&\q{0.76}&   0.24 &200k \\ 
Scene 03  &   24.30 &   0.77 &   0.14 &172k &   24.56 &   0.76 &\q{0.12}&133k &   24.18 &   0.77 &   0.13 &128k &\q{26.57}&\q{0.88}&   0.13 &148k \\ 
Scene 04  &   24.37 &   0.65 &   0.15 &184k &   24.32 &   0.64 &\q{0.12}&162k &   24.37 &   0.67 &   0.13 &140k &\q{25.57}&\q{0.76}&   0.21 &245k \\ 
Scene 05  &\q{20.76}&   0.50 &   0.37 &249k &   19.80 &   0.46 &   0.29 &155k &   20.20 &   0.50 &\q{0.28}&146k &   20.06 &\q{0.56}&   0.30 &329k \\ 
Scene 06  &   22.68 &   0.64 &   0.17 &173k &   23.08 &   0.67 &\q{0.12}&177k &   23.58 &   0.70 &   0.12 &169k &\q{25.64}&\q{0.83}&   0.16 &179k \\ 
\bottomrule
\end{tabular}
\caption{Quantitative comparisons between VAD-GS and other SoTA approaches on the nuScenes dataset. ``\#G'' denotes the number of Gaussian primitives. 
Scene 02 is excluded due to the stationary ego vehicle, which provides no additional supporting views. 
Scenes 07–09 are also omitted, as their extreme nighttime illumination conditions fall beyond the scope of this study. More details are in the supplement.
}
\label{table2.nuscenes}
\end{table*}

\begin{figure*}[t!]
\centering
\includegraphics[width=1.0\textwidth]{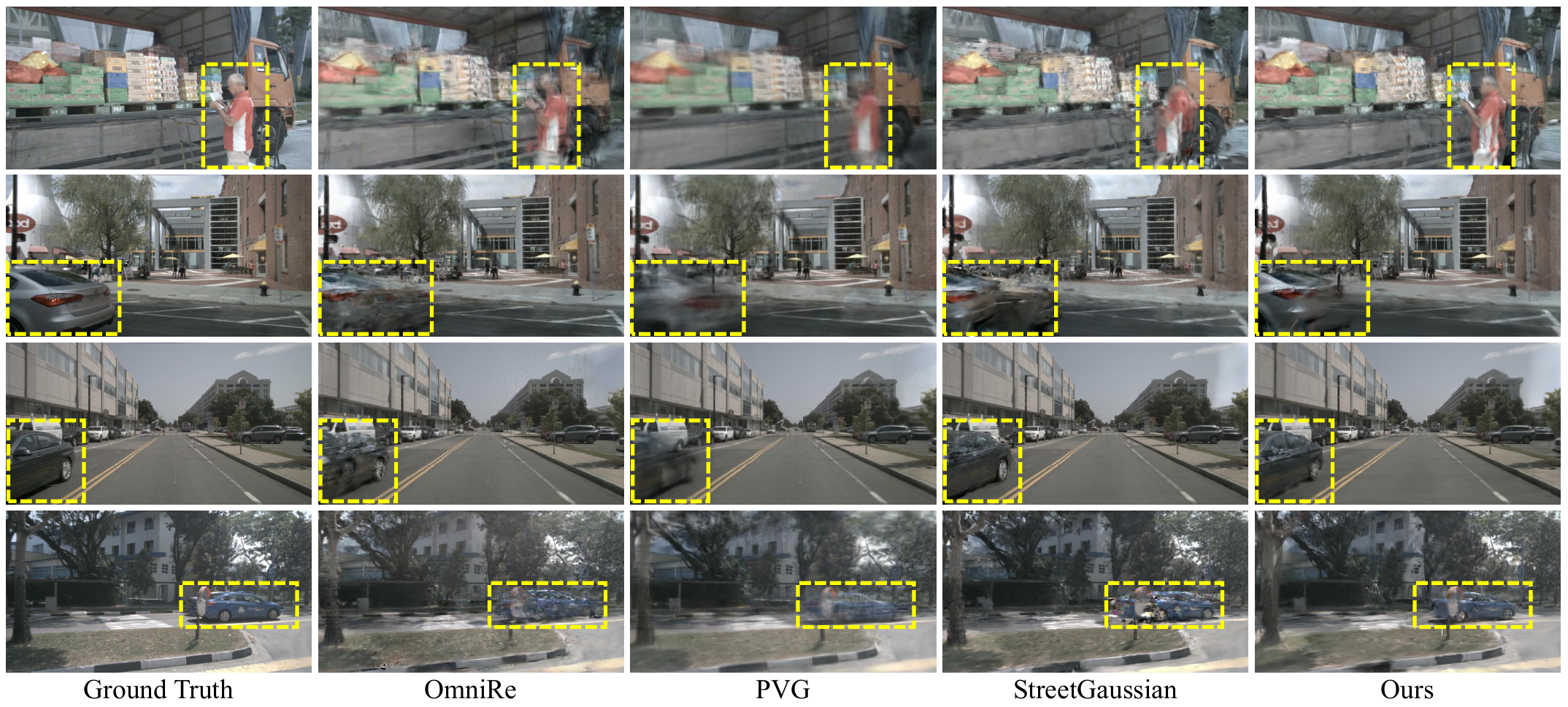}
\caption{
Qualitative comparisons between VAD-GS and other SoTA approaches on the nuScenes dataset.
}
\label{fig5.qualitative}
\vspace{-3mm}
\end{figure*}

\subsection{Datasets}
We conduct extensive experiments on two datasets: Waymo Open \cite{waymo} and nuScenes \cite{caesar2020nuscenes}.
The Waymo Open dataset comprises 1,150 driving scenes recorded in suburban and urban environments. Each frame contains images captured using five cameras and fused point clouds collected using five LiDARs, with an average of 177k points per frame. Following the study \cite{yan2024street}, we select eight sequences, each containing around 100 frames under dynamic traffic conditions.
For the nuScenes dataset, we follow the study \cite{omnire}, which also provides baseline implementations of PVG, OmniRe, and StreetGaussians. 
The data are collected using a 32-beam LiDAR, resulting in significantly sparser point clouds (with an average of 34k points per frame) and more uneven spatial coverage. The increased sparsity poses greater challenges for Gaussian initialization and densification.

\subsection{Implementation Details}
Incomplete structures are identified and re-initialized when either the ratio between the rendered depth and the voxel rasterization result exceeds 1.1 or more than $25\%$ of the pixels belonging to an instance have accumulated opacity below 0.7. The normal and depth loss weights are set to 0.02 and 0.1, respectively, while $\lambda_\text{hard}$ is disabled after 80\% of the training iterations.
Supporting views are selected via maximum-weight $k$-clique optimization, which maximizes the total pairwise similarity among candidate views, where $k$ is set to 4 in practice. 
To alleviate photometric overfitting caused by imbalanced view sampling during optimization, training views are sampled without replacement. Voxel visibility-based densification is performed every five complete sampling cycles. 
Enabled by seamless CUDA integration with the Gaussian training framework, each operation takes approximately 48 ms and is invoked only when missing structures are detected, adding negligible computational overhead.
All experiments are conducted on a single NVIDIA GeForce RTX 4090 GPU.
Please refer to the supplement and video for additional results.

\subsection{Quantitative and Qualitative Results}
We compare VAD-GS with several baseline approaches \cite{3dgs, ost2021neural, wu2023mars, yang2023emernerf, yan2024street, pvg, omnire}. Evaluation metrics, including the peak signal-to-noise ratio (PSNR), the structural similarity index measure (SSIM), and the learned perceptual image patch similarity (LPIPS), are used to quantify models' performance.  As shown in Table \ref{table1.waymo}, VAD-GS consistently outperforms all baseline methods across all evaluation metrics on the Waymo Open dataset. In particular, VAD-GS improves PSNR by $\sim$2.8\% and PSNR* (evaluated on dynamic objects only) by $\sim$3.6\%. These improvements can be primarily attributed to the complemented geometry, which effectively suppresses photometric distortions in erroneously exposed Gaussians. VAD-GS also outperforms the second-best method in terms of SSIM and LPIPS by 0.012 and 0.032, respectively, owing to its more complete and high-fidelity reconstruction of geometry and appearance, which in turn enhances photometric consistency. However, these improvements reflect the potential of our strategy only to a limited extent, as existing baselines report reconstruction results achieved using only the forward-facing camera from the Waymo Open dataset. To ensure fair comparison, we have to adopt the same settings, which restrict the exploitation of cross-camera cues, a key advantage of VAD-GS. 

\begin{figure*}[t!]
\centering
\includegraphics[width=1.0\textwidth]{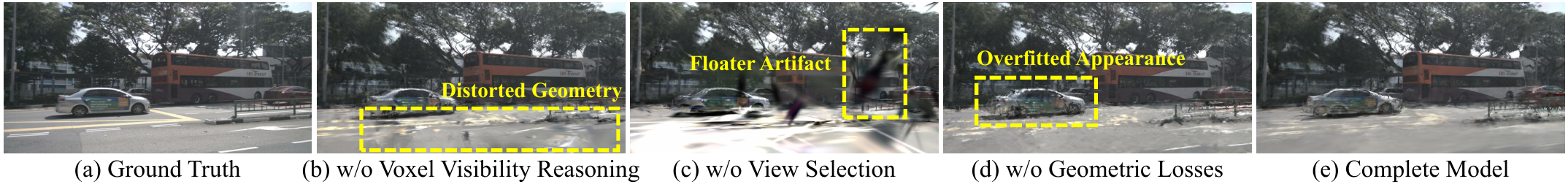} 
\caption{
Qualitative ablation study results.
}
\label{fig6.ablation}
\vspace{-2mm}
\end{figure*}

Completing missing geometry is less critical for the Waymo Open dataset, which provides high-quality point clouds for reliable Gaussian initialization, but becomes essential for the nuScenes dataset due to its significantly sparser LiDAR observations. Given that the difficulty of 3D reconstruction is highly scene-dependent and influenced by factors such as sampling trajectories, occlusions, and dynamic traffic behaviors, we additionally provide per-scene evaluation results on the nuScenes dataset, as shown in Table~\ref{table2.nuscenes} and Fig. \ref{fig5.qualitative}. 
Our method consistently outperforms baseline approaches in terms of SSIM, with improvements exceeding 0.06, and achieves a significant PSNR gain of over 0.78 dB across most scenes, except for Scene 05, where the ego vehicle follows a linear trajectory. This leads to sparse viewpoint sampling and limited overlap across camera views, making it challenging to observe objects from diverse perspectives. 
This results in limited cross-camera cues, making it effectively equivalent to independent single-camera reconstructions.
Our method does not achieve the lowest LPIPS values in several scenes, primarily because the dynamic objects in these scenes are mostly moving pedestrians. Since MVS-based reconstruction methods generally assume object rigidity, their effectiveness degrades when handling deformable or non-rigid objects such as pedestrians. While the number of Gaussian primitives is not a direct indicator of reconstruction quality, it does reflect modeling efficiency to some extent. The slightly higher Gaussian count observed in our method results from targeted densification in underrepresented regions, rather than uncontrolled growth or redundant duplication in well-initialized geometry. 
While quantitative evaluation of structure completion is infeasible without ground truth for unseen trajectories, qualitative visualization from deviated test viewpoints (Fig. \ref{fig1.comparison}) demonstrates the effectiveness of our geometry recovery strategy.

%% file: sec/6_ablation.tex
\begin{table}[t]
\centering
\renewcommand\arraystretch{0.9}
\setlength\tabcolsep{2pt} 
\settablefontnine
    \begin{tabular}{c c c c c}
    \toprule
               Configurations           & PSNR$\uparrow$ & PSNR*$\uparrow$ & SSIM$\uparrow$ & LPIPS$\downarrow$\\
    \hline
    \hline
    w/o voxel visibility reasoning      &    23.79       &     22.75       &    0.753       &    0.215   \\
    w/o view selection                  &    23.92       &     22.83       &    0.757       &    0.212   \\
    w/o geometric losses                & \q{24.59}      &     22.78       &    0.764       & \q{0.194}  \\
    Complete model                      &    24.51       &  \q{23.16}      & \q{0.765}      &    0.199   \\
    \bottomrule
    \end{tabular}
\caption{Quantitative ablation study results.}
\label{table3.ablation}
\vspace{-2mm}
\end{table}

\subsection{Ablation Study}
To validate the efficacy of each component, we conduct a comprehensive ablation study on the nuScenes dataset. Given the interdependence among components (voxel visibility reasoning, view selection, and geometric losses), we train three variants of VAD-GS, each omitting one of these components, and compare their performance against the complete model. Fig.~\ref{fig6.ablation} and Table~\ref{table3.ablation} present the qualitative and quantitative results, respectively. 
The first variant removes the voxel-based visibility reasoning component, which consequently disables all other components. Although photometric-based densification remains active, it fails to accurately recover unreliable geometry. As shown in Fig.~\ref{fig6.ablation}(b), this leads to incorrect gradient updates that distort Gaussian primitives, ultimately causing significant performance degradation. 
The second variant disables diversity-aware view selection and instead relies on fixed consecutive frames for patch matching. 
Although this improves densification in static regions, the recovery of missing geometry remains incomplete. Moreover, the absence of explicit separation between static and dynamic regions causes misleading matches between background and foreground geometries, leading to severe floater artifacts, as shown in Fig.~\ref{fig6.ablation}(c).
The third variant excludes geometric losses from the optimization objective. While it achieves comparable or even slightly better photometric metrics, attributed to the strong reliance on image similarity as the sole supervision signal, which encourages overfitting to visual appearance, this variant introduces noticeable artifacts under large viewpoint deviations, such as the rough vehicle surfaces observed in Fig.~\ref{fig6.ablation}(d).

%% file: sec/7_conclusion.tex
\section{Conclusion and Future Work}
This study presented VAD-GS, a novel 3DGS framework designed to enhance geometry recovery under sparse observations, particularly in dynamic, unbounded urban environments. Unlike prior Gaussian densification methods that exclusively clone or split existing Gaussians, VAD-GS reconstructs new Gaussians by leveraging stereo cues from multiple cameras across different timestamps, effectively recovering missing geometry for both static and dynamic objects. 
The framework explicitly models the view-dependent voxel visibility, enabling occlusion reasoning among scene instances and identification of regions requiring reconstruction. It then strategically selects supporting views based on a newly defined diversity score and generates complementary point clouds that satisfy multi-view consistency for objects even in motion. Extensive experiments on public datasets demonstrate the superiority. Despite the competitive performance in geometry completion, certain limitations persist. As a trade-off for relaxing static MVS assumptions in dynamic settings, the proposed method imposes a rigidity constraint. Consequently, in the absence of rigidity, densification for deformable objects degrades to a standard gradient-based strategy. Extending the approach to handle non-rigid motion is non-trivial and is left for future~work.

%% file: sec/8_acknowledge.tex
\section*{Acknowledgment}
This research was supported by the National Natural Science Foundation of China under Grants 62473288, 62233013, 62388101, and 62333017, the National Key Laboratory of Human-Machine Hybrid Augmented Intelligence, Xi'an Jiaotong University (No. HMHAI-202406), NIO University Programme (NIO UP), the Fundamental Research Funds for the Central Universities, and the Xiaomi Young Talents Program. \\

%% file: sec/X_suppl.tex
\clearpage
\setcounter{page}{1}
\maketitlesupplementary

\clr{
}
\section{Multi-Camera Cross-Frame Views}
\begin{figure}[t!] 
\centering
\includegraphics[width=1.0\columnwidth]{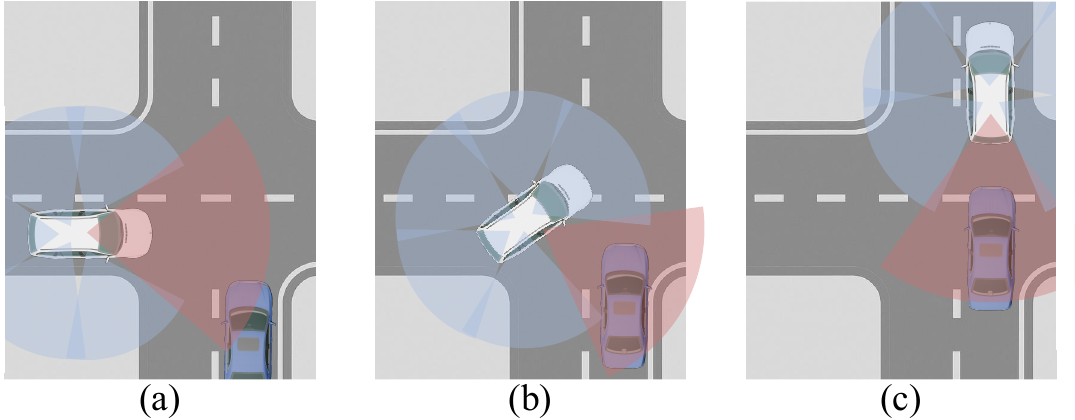} 
\caption{
\textbf{An illustration of multi-camera, cross-frame views.}
For both static and dynamic objects, informative observation views are typically captured by different cameras at different timestamps.
}
\label{fig_a1.frustum}
\end{figure}

As shown in Fig.~\ref{fig_a1.frustum}, the outward-facing multi-camera views have limited overlaps. Prior methods such as \cite{3dgs} typically treat all views indiscriminately during Gaussian training, regardless of their spatial or temporal differences. Nonetheless, structural complexity varies significantly across regions, necessitating a selective reconstruction strategy that prioritizes critical objects over trivial or redundant structures.
Object-centric reconstruction strategies generally assume sufficient overlap among views within a bounded range and minimal interference from unrelated perspectives. However, this assumption breaks down in dynamic, unbounded urban scenes.
The failure case illustrated in Fig. \ref{fig_a1.frustum} suggests that observations from the same camera fail to continuously capture a moving target vehicle.

\subsection{Visibility Reasoning}
Visibility determination, also known as hidden surface removal (HSR) or occlusion culling (OC), which identifies visible surfaces from a given viewpoint, has long been a central topic in computer graphics \cite{cohen2003survey}. 
Among numerous HSR algorithms, z-buffering is usually the choice due to its simplicity and efficient hardware implementation.
In contrast, Gaussian splatting renders pixels by alpha-blending all primitives along each viewing ray rather than explicitly enforcing occlusion. 
As points cannot occlude one another, no primitive is truly hidden, as illustrated in Fig.~\ref{fig_a0.point_projection} in the supplement. 
With sufficient viewing directions, primitives on visible surfaces may eventually become opaque, which implicitly recovers occlusion. 
However, when initialization is incomplete and views are limited, ambiguity arises: observed appearance is simply mapped onto whichever primitives are projected to the image plane, regardless of whether the corresponding surface exists.
Since incomplete geometry would inevitably mislead optimization, densification strategies have to take visibility reasoning into consideration to assess surface completeness before applying further updates.
Most existing Gaussian splatting methods take COLMAP \cite{colmap} point clouds directly as Gaussian centers and overlook their associated visibility information.  
Yet COLMAP inherently encodes rich visibility cues.  
Specifically, the ``TRACK'' table records the source views in which each 3D point is observed and successfully triangulated.  
This allows us to exploit the following two real-world visibility observations in a more effective way:  
(1) Each 3D point lies on the first surface intersected by the pixel rays from all source views, implying that the line of sight between the object and each viewpoint is occlusion-free. 
(2) The local structure surrounding each 3D point is visible in its associated source views, thereby providing reliable supporting-view candidates for object reconstruction.

\begin{figure}[t!] 
\centering
\includegraphics[width=1.0\columnwidth]{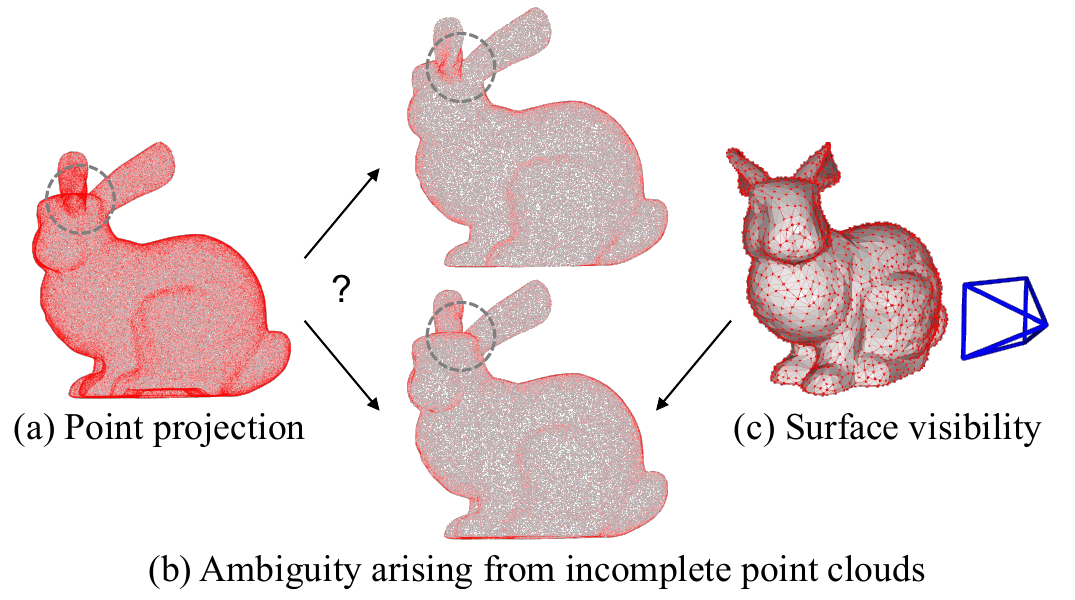} 
\caption{
\textbf{Is the rabbit facing forward or backward?}
(a) Points do not occlude one another, introducing appearance and geometry ambiguity. 
(b) Incomplete initialization may force the model to map front-side appearance onto back-side primitives, which are then incorrectly optimized and become distorted. 
(c) Identifying and completing missing structures requires densification with explicit visibility reasoning.
}
\label{fig_a0.point_projection}
\end{figure}

\subsection{View Selection}
The diversity score $s$ introduced in the main paper quantifies the geometric dissimilarity between a pair of views. 
However, selecting an informative subset of supporting views for reconstruction requires more than simply maximizing diversity between view pairs, ensuring that the subset is collectively informative and non-redundant. Moreover, as the same reference object may appear repeatedly during training, a deterministic selection based solely on diversity may lead to overfitting or limited generalization. To address this issue, we propose to sample views via:
\begin{equation}
\begin{aligned}
    \max_{\mathcal{V}_s \subset \mathcal{V}_c }
    \sum_{v_i \in \mathcal{V}_s} s_{iR} \xi_{iR} + \lambda \sum_{\{v_i, v_j\} \subset \mathcal{V}_s} s_{ij} \xi_{ij},& \\ 
    |\mathcal{V}_s| = k, \quad \xi \sim \mathcal{N}(1, \epsilon),&
\end{aligned}
\end{equation}
where $\mathcal{V}_c$ denotes the full set of all candidate views, $\mathcal{V}_s$ represents the selected subset containing $k$ supporting views, $s_{iR}$ denotes the diversity score between each pair of candidate view and the reference view, $s_{ij}$ represent the diversity score among views within the subset, and $\epsilon$ represents a noise term introduced to encourage sampling diversity. This randomized selection strategy ensures relevance to the reference view while avoiding deterministic bias, resulting in a diverse yet non-redundant subset of supporting views.

\section{Additional Experiments}

\begin{figure*}[t!] 
\centering
\includegraphics[width=1.0\textwidth]{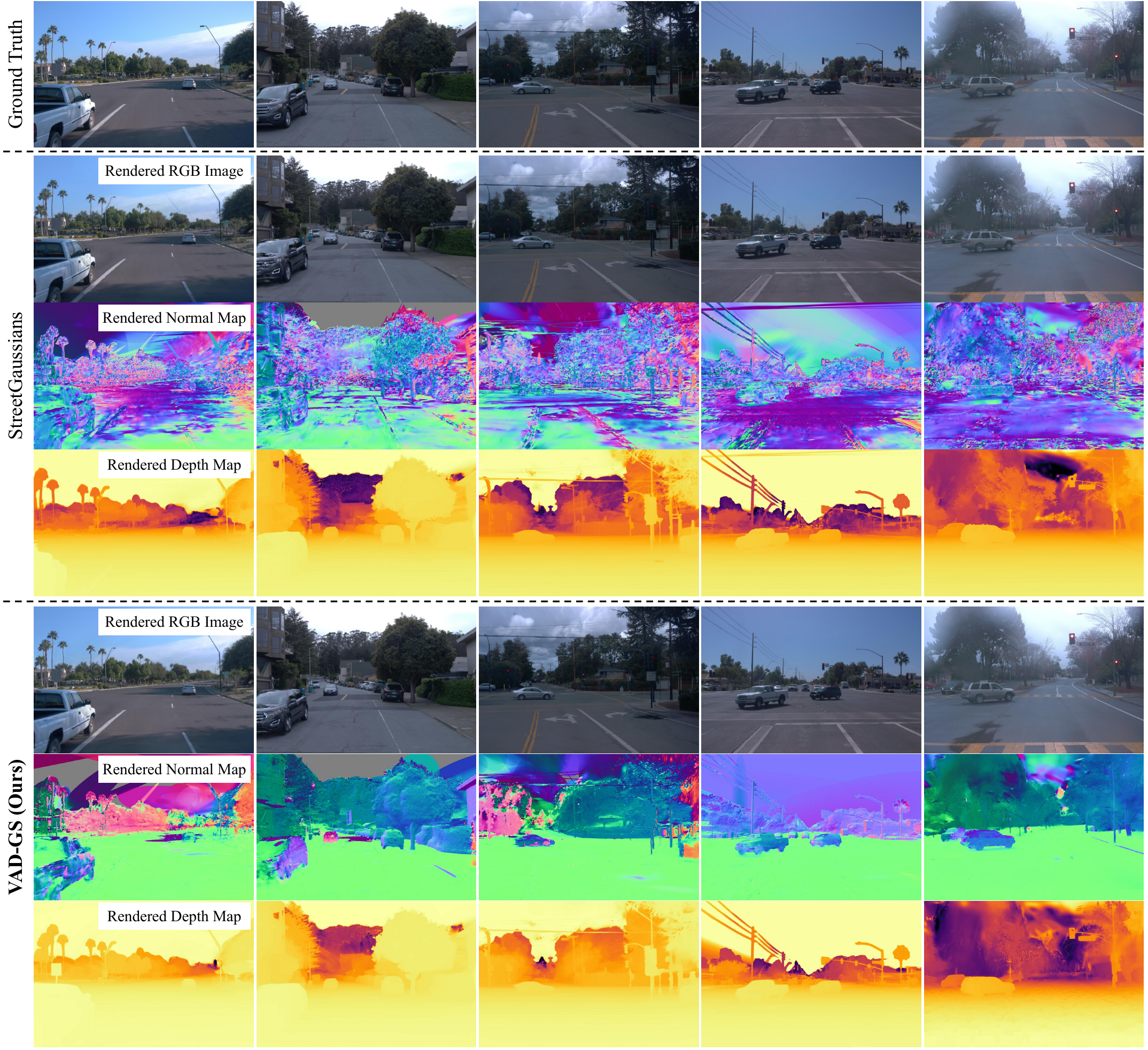} 
\caption{
\textbf{Additional qualitative results on the Waymo Open dataset.} 
Due to the single-camera configuration, test views captured by the forward-facing camera exhibit substantial overlap with the training views. While all methods achieve high-fidelity rendering results under this setting, such performance may not reliably indicate the quality of the underlying geometry.
}
\label{fig_a2.waymo_qualitative}
\end{figure*}

\subsection{Experimental Details} 
While many 3DGS methods adopt similar train/test splitting strategies, the specific details on these splits remain ambiguous for urban driving scenes. For example, statements such as ``randomly select every \textit{n}-th image of different cameras''can be interpreted in multiple ways: either as discarding specific frames with all associated camera views, or as selectively omitting individual views while retaining the full sequence of frames. Moreover, such random sampling schemes are misaligned with the practical goal of novel view synthesis, which aims to render intermediate views between consecutive video frames captured by multi-camera systems mounted on a moving vehicle.

While both strategies remove the same number of views, randomly selecting individual test views results in more uniform frustum coverage and visually cleaner outputs. However, this approach exploits temporal redundancy and overlooks the realistic constraint that multi-camera views are typically available or missing as a complete observation. In contrast, removing all views at specific timestamps significantly reduces scene coverage and degrades visual quality, particularly when the vehicle is moving rapidly. Despite being more challenging, this setting better reflects real-world deployment constraints and more effectively evaluates the model's generalizability.

Specifically, we select every fourth frame along with all associated camera views to construct the test set.  As a result, spatial observations are entirely unavailable for approximately 25\% of the ego vehicle poses. This setting poses significant challenges for models that rely on multi-view consistency or temporal cues, and serves as a rigorous benchmark for evaluating reconstruction robustness under sparse observational conditions.

\begin{figure*}[t!] 
\centering
\includegraphics[width=1.0\textwidth]{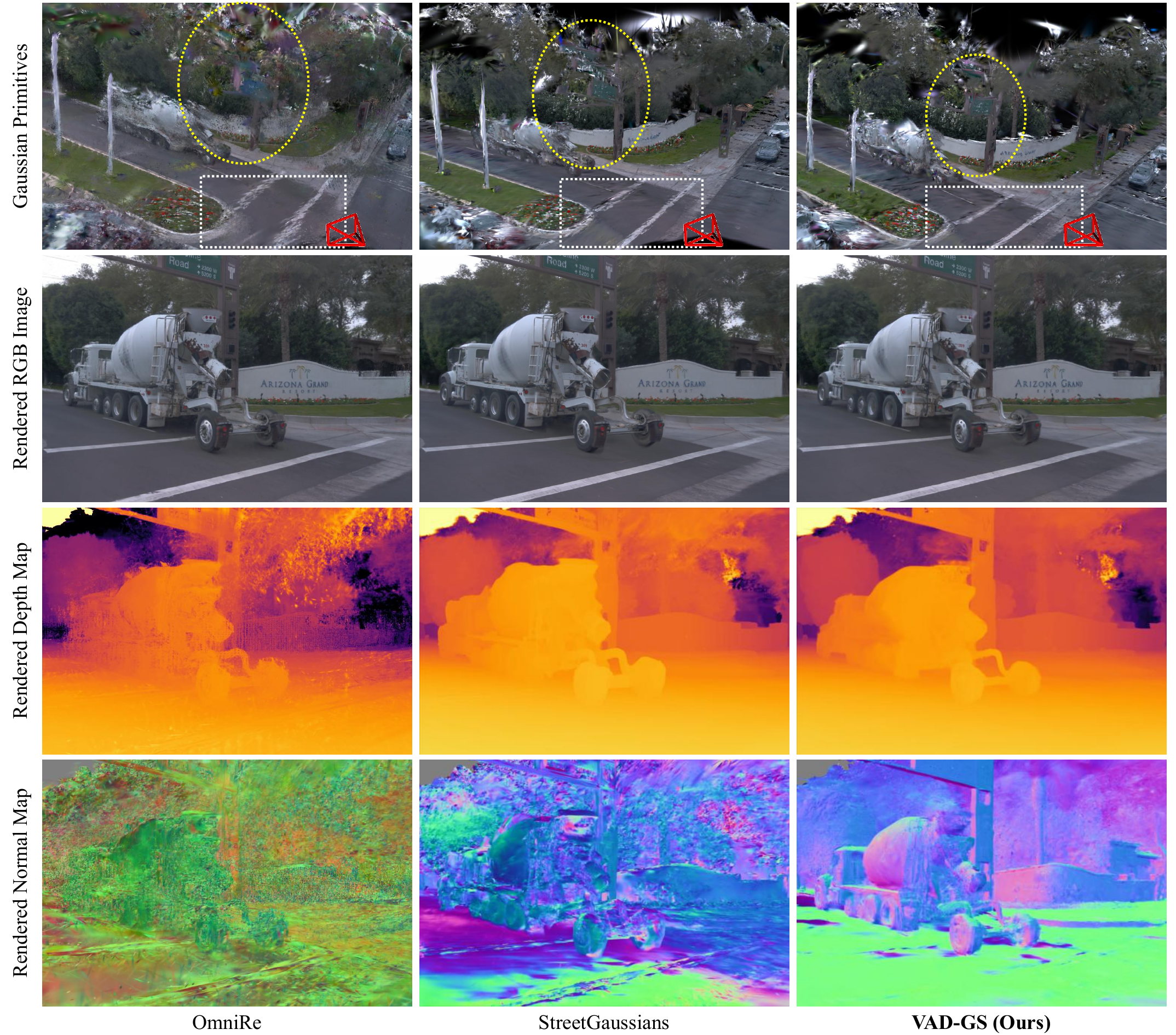} 
\caption{
\textbf{Qualitative comparison between VAD-GS and other SoTA methods on the Waymo Open dataset when a multi-camera configuration is used}. }
\label{fig_a3.waymo_visualize}
\end{figure*}

\begin{figure*}[t!] 
\centering
\includegraphics[width=1.0\textwidth]{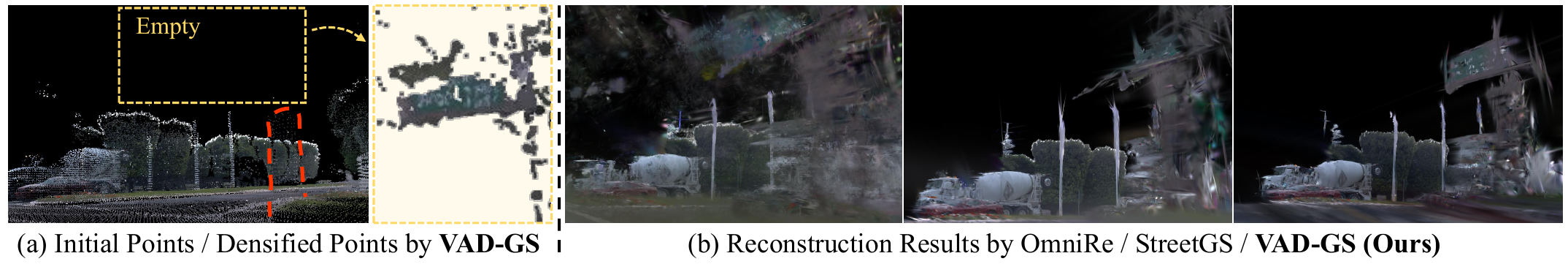} 
\caption{
\textbf{Qualitative comparison of reconstructed scene geometry}. 
Revisiting the Waymo traffic sign example given in the paper, the LiDAR and SfM points are not only noisy but \textbf{missing}, with only a few points available within the \clr{red} dashed area. In this extremely challenging case, the exact contribution of our densification lies in recovering points in the \cly{yellow} box.
}
\label{fig_a7.necessity}
\end{figure*}

\begin{figure*}[t!] 
\centering
\includegraphics[width=1.0\textwidth]{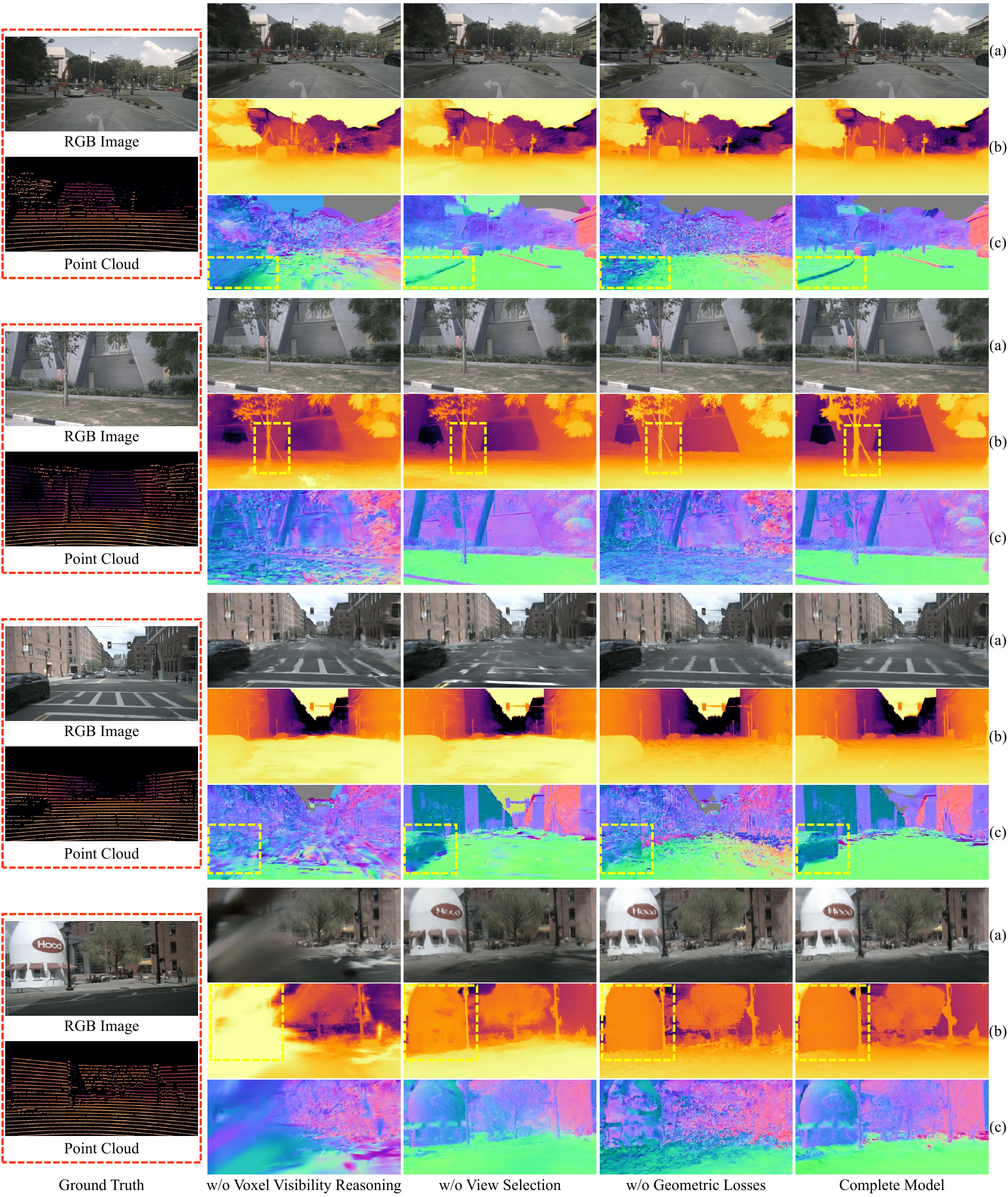} 
\caption{
\textbf{Additional qualitative ablation study results on the nuScenes dataset.} 
The rendered RGB images, depth maps, and normal maps are visualized in (a), (b), and (c), respectively.
}
\label{fig_a5.add_ablation}
\end{figure*}

\subsection{Additional Qualitative Comparisons}
In this supplement, we provide additional comparative results against recent methods on large-scale driving scenes. Due to the page limitation, qualitative results on the Waymo Open dataset \cite{waymo} are provided in Fig.~\ref{fig_a2.waymo_qualitative}. For fair comparison, we adopt the validation configuration of StreetGaussians \cite{yan2024street} and use only a single forward-facing camera. This setup simplifies view-dependent appearance and geometry consistency constraints, as the forward-facing view undergoes relatively minor temporal changes. However, it inherently limits the acquisition of novel information and significantly reduces overall scene coverage. These minimal inter-frame variations result in highly similar and redundant observations, which can provide limited geometric diversity for triangulation or multi-view spatial-consistency reasoning, thus failing to fully unleash the potential of visibility-aware densification for complete geometry reconstruction. Consequently, high-fidelity rendering quality may not indicate accurate scene geometry recovery, but rather reflect overfitting to specific image observations.

To further demonstrate the high quality of our scene reconstruction, we present an additional example in Fig.~\ref{fig_a3.waymo_visualize}. This comparison is performed by adopting a multi-camera configuration that utilizes cameras 0, 1, and 2 from the Waymo Open dataset. Although all methods achieve comparable rendering quality, the underlying geometry differs significantly. The traffic sign, highlighted by yellow circles, lies outside the LiDAR scanning range and is only partially visible from a limited number of viewpoints. In OmniRe \cite{omnire}, the sign is reconstructed as a set of scattered and unstructured Gaussians, indicating overfitting to appearance cues in the absence of reliable geometric constraints.
As for StreetGaussians, the sign appears fragmented and discontinuous, with Gaussians erroneously updated to positions between the sign and the background trees. These artifacts stem from missing Gaussians caused by incomplete initialization, which in turn lead to erroneous gradient propagation toward trees that should be occluded. The misdirected gradients distort the initial Gaussians representing the leaves, altering their color, position, and shape, and unnaturally pull them toward the sign, ultimately resulting in fragmented and misaligned geometry.

Benefiting from visibility reasoning, view selection, and MVS-based reconstruction, VAD-GS densifies Gaussians beyond conventional photometric-based splitting and cloning strategies, greatly alleviating issues related to incomplete or distorted geometry. Notably, VAD-GS accurately recovers the planar structure of the traffic sign, with only minor artifacts at the top border due to limited observations. Additionally, the road surface, highlighted by the white box, demonstrates a more geometrically consistent reconstruction compared to other approaches.

\subsection{Additional Ablation Studies}
In this supplementary material, we also present additional qualitative ablation study results, including rendered RGB images, depth maps, and normal maps, to further demonstrate the effectiveness of each module in VAD-GS. As shown in Fig.~\ref{fig_a5.add_ablation}, the sparse point clouds provide limited surface coverage. Each LiDAR scan line in the ground-truth point typically contributes only two or three points to thin structures such as tree trunks or utility poles. Additionally, due to the limited scanning angle and sparse sampling intervals, the resulting point cloud distribution exhibits substantial gaps and covers only a narrow field of view. These limitations pose significant challenges for capturing complete geometry, particularly for large and distant surfaces such as buildings and walls. 

Furthermore, we select several challenging test views to more clearly demonstrate the contribution of each component. A common issue during densification is the emergence of floaters, where Gaussians become misaligned with the actual scene geometry. While most floaters are often naturally pruned or corrected when they appear in regions well-covered by training views, they tend to persist in sparsely observed areas. In selected test views where these floaters are prominent, our geometric loss effectively penalizes them, encouraging alignment with the correct underlying surfaces. This process significantly improves the final surface quality and substantially reduces visual artifacts. Moreover, objects that are only transiently visible, such as moving vehicles or structures primarily observed from side views, often suffer from sparse observations. Our view selection and MVS-based reconstruction modules improve the instance-level fidelity in these challenging regions, including dynamic vehicles, small trees, and complex landmarks such as the bottle-shaped building. 

\section{Failure Cases and Limitations}
Despite achieving high-fidelity performance, VAD-GS still exhibits several known limitations. 
The primary challenge lies in its inability to effectively handle deformable objects, such as pedestrians. Given that our objective is to recover geometry in complex urban scenes, the presence of walking pedestrians is inevitable.  Nonetheless, these non-rigid objects violate the rigidity assumption required by MVS-based reconstruction. Future work will explore the integration of state-of-the-art Gaussian-based deformable object modeling approaches, such as 4DGS~\cite{4dgs} and SC-GS~\cite{scgs}, to address this issue.

Second, our method assumes locally consistent visibility among neighboring points. While this assumption enables effective occlusion modeling and supports continuous surface reconstruction, it may fail in extreme cases involving complex structures such as wire fences or glass surfaces. These structures often reflect LiDAR beams, producing dense point clouds that resemble those from regular surfaces. Nevertheless, the simultaneously captured images may reveal background objects without occlusion, leading to discrepancies between geometric and visual observations. Accurately and efficiently modeling occlusion relationships in such challenging and visually ambiguous regions remains an important direction for future research.
